%% file: workshop_paper.tex
\def\arxiv{1}

\ifdefined\arxiv
  \documentclass[sigconf,nonacm]{acmart}   
\else
  \documentclass[sigconf,review]{acmart}  
\fi

\setcopyright{acmlicensed}
\copyrightyear{2027}
\acmYear{2027}
\acmDOI{XXXXXXX.XXXXXXX}   
\acmConference[KDD '27]{Proceedings of the 33rd ACM SIGKDD Conference on
Knowledge Discovery and Data Mining}{August 2027}{San Jose, CA, USA}
\acmISBN{978-1-4503-XXXX-X/2027/08}  

\usepackage{graphicx}
\usepackage{natbib}
\usepackage{hyperref}
\hypersetup{
    colorlinks=true,
    linkcolor=black,
    filecolor=black,
    urlcolor=cyan,
    citecolor=black
}
\usepackage{amsmath}
\usepackage{booktabs}
\usepackage{amsfonts}
\usepackage{nicefrac}
\usepackage{microtype}
\usepackage{xcolor}
\usepackage{float}
\usepackage{tikz}
\usetikzlibrary{positioning, arrows.meta, fit, backgrounds, calc}
\definecolor{okblue}{HTML}{0072B2}
\definecolor{okverm}{HTML}{D55E00}

\usepackage{tcolorbox}
\tcbuselibrary{listings, breakable}

\newtcblisting[auto counter]{promptbox}[2][]{
  listing only, breakable,
  colback=gray!5, colframe=black!60,
  title={Prompt~\thetcbcounter: #2}, fonttitle=\bfseries,
  size=small,
  listing options={basicstyle=\small\ttfamily, breaklines=true,
                   columns=fullflexible, keepspaces=true,breakindent=0pt,
breakautoindent=false},
  #1
}

\begin{document}
\title{Offline-to-Online Creative Optimization with Generative Models and Adaptive Testing}

\author{Kevin Lee}
\affiliation{
\institution{University of Michigan} \country{USA}
}
\email{kvnlee@umich.edu}

\author{Benjamin Letham}
\affiliation{
\institution{Meta}
\country{USA}
}
\email{bletham@meta.com}

\author{Zhiyuan Jerry Lin}
\affiliation{
\institution{Meta}
\country{USA}
}\email{zylin@meta.com}

\author{Elodie Samson}
\affiliation{
\institution{Meta}
\country{USA}
}\email{elosamson@meta.com}

\author{Eric Onofrey}
\affiliation{
\institution{Meta}
\country{USA}
}\email{eonofrey@meta.com}

\author{Poppy Zhang}
\affiliation{
\institution{Meta}
\country{USA}
}\email{poppyzhang@meta.com}

\author{Shawndra Hill}
\affiliation{
\institution{Meta}
\country{USA}
}\email{shawndrahill@meta.com}

\author{Eytan Bakshy}
\affiliation{
\institution{Meta}
\country{USA}
}\email{ebakshy@meta.com}

\renewcommand{\shortauthors}{Lee et al.}


\begin{abstract}
Ad creative optimization is increasingly constrained by evaluation rather than generation. Generative models can produce many plausible creatives, but reliable evaluation requires online experiments, in which only a limited slate can be tested. We study how to use data from historical A/B tests to generate and select the candidates in that slate. We developed and deployed a performance-driven offline-to-online workflow that guides creative generation with a predictive model as an inference-time critic. In the offline phase, we use a predictive model trained on historical experiments to rank and refine variants created by a generative model. A final test slate is then deployed in an online adaptive experiment. In a 50-arm field experiment, we found that the best creative generated with this method yielded 45.1\% higher engagement than the best human-authored creative. Two additional experiments showed the same upper-tail pattern, with lifts of 46.7\% and 36.2\%. We found that despite the predictive model being too noisy to directly identify the best creative offline, it effectively guides the generative model toward creating strong candidates that can be efficiently evaluated in an adaptive experiment. The results suggest a design principle for creative optimization with generative models: use predictive models to guide generation of a slate to test, judge the slate by whether it contains high-performing candidates at a feasible test size, and use adaptive experiments to select among candidates while limiting traffic lost to weak arms.

\end{abstract}

\maketitle

\section{Introduction}

Generative AI is transforming advertising creative generation from manual copywriting to high-dimensional semantic search: given an infinite space of possible messages, which specific message will maximize user engagement and conversion?  Traditionally, firms ask human experts to write a small number of candidate creatives and use judgment, possibly paired with small-scale A/B testing, to choose from among them. Generative models have made the first step much cheaper. A marketer can now produce practically endless variants of creatives (ad copies, push notification messages, marketing emails, etc.) with little effort. However, this new paradigm has only shifted the bottleneck from generation to evaluation. The marketer still must identify the best creative, and a large set of candidate variants makes the evaluation and selection problem much more difficult. The gold-standard method for evaluating ad creatives is to run field experiments with online traffic, where noise levels are generally high and only a limited slate can be tested. Every additional candidate consumes traffic, delays selection, and potentially exposes users to weaker arms.

This paper studies how to use data from historical A/B tests together with an LLM to generate and select the creative candidates to evaluate. In digital advertising, firms often have data from past randomized experiments that measured which creatives performed better in prior campaigns. Typical workflows with predictive models trained on these data usually involve scoring creatives and deploying the one predicted to perform best. However, ad creatives are high-dimensional, campaign context can change dramatically, the collected engagement outcomes are noisy, and the generated candidates can also move outside the support of past experiments. As a result, the predictive model is often not accurate enough to directly choose the winning creative fully offline. Instead, we show that offline predictive models can still be useful when used to guide the generation toward candidates worth testing. The key insight of our approach is simple:

\begin{quote}
\textbf{\textit{The predictive model does not need to pick the winner; it needs to help generate a slate that contains winners.}}
\end{quote}
We implement this idea in an offline-to-online workflow with two stages:

\begin{itemize}
    \item \textit{Candidate generation}: In the offline stage, a frozen generative model starts from human-written seed creatives and proposes variants. A predictive model trained on historical A/B tests ranks those variants. The ranking then guides the generative model in one refinement step, producing additional candidates. The pooled candidates are re-ranked to form the online test slate. The predictive model is therefore used as an inference-time critic for generation, not as a deployment rule.
    \item \textit{Adaptive screening}: In the online stage, the resulting slate is evaluated in an adaptive experiment. The experiment uses real user feedback to select from the included candidates. Adaptive allocation can rapidly identify the best candidate while mitigating the effect of false positives by shifting traffic away from weak arms during the test. The two stages are complementary: the offline stage generates a slate likely to contain high-performing candidates (high recall), and the online stage supplies the precision needed to choose from among them.
\end{itemize}

The predictive model is judged on whether the slate it helps generate contains high-performing candidates at a size the experiment can afford. Including weak candidates reduces overall performance because they consume traffic, but excluding strong candidates is worse because the online experiment cannot recover candidates that never enter the slate. We therefore evaluate the workflow using three metrics: candidate-set recall for the generated slate, selected-arm quality for the final decision, and regret during online testing.

Our system was deployed for creative optimization at Meta, and we describe here three distinct large-scale optimizations that have used the system. In our main 50-arm experiment, the six top-performing creatives were all AI-refined, and the best AI-refined creative achieved 45.1\% higher engagement than the best human-authored creative. The predictive model is not reliable enough for direct deployment: in that experiment its highest-scored creative finished 43rd of 50. However, a top-30 ranked slate contains five of the six creatives that outperformed the best human creative. Simulations calibrated to the empirical results show that adaptive allocation preserves selected-arm quality while reducing the opportunity cost of screening weak candidates.

\textit{Contributions.}
Our work provides guidance on how creative optimization can be improved by incorporating offline data and predictive models into the generation process. First, we show how predictive models trained on historical A/B tests can guide the generation of candidates for a new online experiment, not merely rank candidates after they have been produced.
We demonstrate that historical experimental data can shape which creatives are generated and tested, without retraining the generator or treating the predictive model as a deployment rule.

Second, we provide a criterion for evaluating prediction models in the new era of AI generation: the model should not be used to pick the top-1 creative as it has been used traditionally.  Rather, it should help generate sets of creatives containing high-performing candidates at the size the online experiment can afford, and thus be evaluated on candidate-set recall.

Lastly, we report results from three optimizations run on our deployed system. AI-refined creatives concentrate in the upper tail of the tested slate. At the same time, the predictive model produces enough false positives that it should not be used as a deployment rule. Adaptive online testing turns the high-recall slate into a high-performing final decision while reducing traffic lost to weak candidates.

\section{Related Work}
Our work connects to offline-to-online optimization, two-stage retrieval systems, adaptive experimentation, and recent work on generative AI for marketing and text optimization.

\textit{Offline-to-online optimization.}
The transition from offline data to online deployment is a central problem in reinforcement learning and decision making. Standard offline RL methods, such as Conservative Q-Learning \citep{kumar2020conservative} and Batch-Constrained Q-learning \citep{fujimoto2019off}, focus on mitigating distributional shift by penalizing actions outside the training distribution. This problem is particularly challenging in high-dimensional spaces like text, where the action space is combinatorial and vastly larger than the training set \citep{levine2020offline}. Our approach shifts the focus from direct action selection to candidate set generation, related to hybrid frameworks that use offline data to warm-start or supplement online exploration \citep{letham2019bayesian,nair2020awac}. While traditional offline-to-online work typically assumes a fixed action space, we use the offline model within a generative process to construct a dynamic action space tailored for subsequent online screening.

\textit{Two-stage retrieval and ranking.}
Our conceptual framework of using an offline model as a high-recall generator rather than a high-precision selector is related to the architecture of industrial recommender systems. Classic two-stage systems employ a \textit{retrieval} stage for high-recall candidate generation followed by a \textit{ranking} stage for high-precision scoring \citep{covington2016deep}. This design is used in industrial-scale settings where the space of items to rank is much larger than what could be directly evaluated with a high-precision model. Creative generation faces a similar bottleneck, in that the cardinality of the set of possible creatives is far too high for direct evaluation via online testing. We argue that the best use of the offline model is as a first stage for high-recall candidate generation \citep{karpukhin2020dense} and to ensure that top performers are present in the generated slate, rather than trying to accurately predict their rank.

\textit{Adaptive experimentation and batched bandits.}
The online component of our work uses multi-armed bandits and batched adaptive experimentation. Fully sequential bandits are often infeasible in digital advertising because decisions are made in batches and because reporting, delivery, and system constraints introduce latency. Batched bandit methods address this practical regime \citep{perchet2016batched,schwartz2017customer}. Our workflow uses a batched adaptive design to screen AI-generated candidates, building on the literature on best-arm identification \citep{even2006action, jamieson2014lil} and fixed-budget experimentation \citep{audibert2010best, karnin2013almost}, where the goal is to identify the top-performing treatment while minimizing the regret incurred during the experiment.

\textit{LLMs and marketing.}
Recent research in marketing addresses how language models can generate, improve, or evaluate customer-facing copy. \citet{reisenbichler2026sponsored} develop a tailored LLM application for sponsored search advertising and evaluate it in empirical advertising settings. \citet{angelopoulos2024causal} use within-A/B-test comparisons to train a model to improve marketing copy and introduce the Bradley--Terry ranking component that we use here to score generated candidates. \citet{jiang2025adllama} post-train a generative ad-text model using historical performance feedback and evaluate it in a large-scale online experiment. \citet{ye2025lola} integrate LLM-based predictions with online learning for content experiments. These predictions are used to initialize allocation probabilities over a fixed, pre-existing content set using LLM-informed priors, whereas we generate the content set. We share the broad idea that historical performance data can improve generated marketing text, but our deployment architecture differs: the predictive ranking model is used as an inference-time critic to steer slate construction, and a new batched online experiment makes the final selection.

\textit{Inference-time search in language space.}
Our generation procedure is also related to inference-time text optimization \citep{agrawal2025gepa}. TextGrad-style methods use textual critiques or local edits to improve candidates without updating model weights \citep{yuksekgonul2024textgrad}. TextBO formalizes a closely related Best-of-$N$ plus textual-gradient search pattern in language space, showing that the procedure is able to optimize a reward function \citep{kang2025textbo}. We apply a similar inference-time pattern, but use the signal from an offline predictive model to construct candidates for online testing. Our approach also connects to the emerging ``LLM-in-the-loop'' optimization paradigms where the generative model explores the space while an external verifier or experiment provides the ground truth \citep{yang2024large}. In our system, the score used to guide inference-time search is a Bradley--Terry ranker trained from randomized experiments, and the final arbiter is an online field experiment rather than an LLM judge or purely offline evaluator.


\section{Problem Setup}

\subsection{Setting}

Let \(x \in \mathcal X\) denote the context of an ad campaign, e.g. audience, channel, objective, and timing. Let \(y \in \mathcal Y\) denote the content of an ad creative, e.g. ad text or email copy. In context \(x\), creative \(y\) has expected engagement
\[
\mu(x,y) = \mathbb{E}[r \mid x,y],
\]
where \(r \in \{0,1\}\) is an engagement indicator, such as an open or a click.

We observe results from historical randomized experiments. In each experiment, several creatives are shown under random assignment in a common context, so differences in realized engagement identify which creative performed better in that context. From each experiment we extract ordered comparisons \(y_A \succ y_B\), read as ``\(y_A\) outperformed \(y_B\).'' Aggregated across experiments, these within-experiment comparisons are the training signal for the ranking model.

We do not treat cardinal engagement rates as directly comparable across experiments. Baseline performance can differ because of audience composition, seasonality, placement, campaign objective, and other context-level factors. The ranker is therefore trained from ordinal comparisons within randomized experiments rather than from raw cross-campaign outcome levels.

\begin{figure*}[t]
\centering
\begin{tikzpicture}[
  font=\small,
  box/.style={draw=black!55, rounded corners=2pt, align=center, fill=white,
              inner sep=4pt, minimum height=1.1cm, text width=2.6cm},
  flow/.style={-{Stealth[length=2.4mm]}, semithick, black!60},
  plabel/.style={font=\small\bfseries, text=black!75},
  panel/.style={draw=black!30, rounded corners=4pt},
]
\node[box] (seed) {Seed creative};
\node[box, right=10mm of seed] (gen)  {Generate\\variants};
\node[box, right=10mm of gen]  (rr)   {Rank \&\\refine};
\node[box, right=10mm of rr]   (test) {Adaptive online\\test};
\draw[flow] (seed) -- (gen);
\draw[flow] (gen)  -- (rr);
\draw[flow] (rr)   -- (test);
\node[box, above=14mm of rr] (ranker) {Predictive ranking\\model};
\node[box, left=14mm of ranker] (data) {Historical\\randomized\\experiments};
\draw[flow] (data) -- (ranker);
\draw[flow, dashed] (ranker) -- (rr);
\draw[flow] ($(rr.south)+(-3.5mm,0)$) to[out=-90,in=-90,looseness=1.3]
      node[midway,below=0.2mm,font=\scriptsize,text=black!60]{} ($(rr.south)+(3.5mm,0)$);
\node[inner sep=0] (loopanchor) at ($(rr.south)+(0,-4mm)$) {};
\node[plabel, above=1.5mm of data.north west, anchor=south west] (ttrain) {Training};
\node[plabel, above=1.5mm of seed.north west, anchor=south west] (tinfer) {Deployment};
\begin{scope}[on background layer]
  \node[panel, fill=okblue!5, fit=(ttrain)(data)(ranker)] {};
  \node[panel, fill=okverm!6, fit=(tinfer)(seed)(gen)(rr)(test)(loopanchor)] {};
\end{scope}
\end{tikzpicture}
\caption{Training (top): data from previous experiments trains a ranking model. At deployment (bottom), seed creatives are expanded into candidate variants, iteratively ranked and refined by the predictive model, then screened by an adaptive online test that selects the final creative.}
\label{fig:workflow}
\end{figure*}
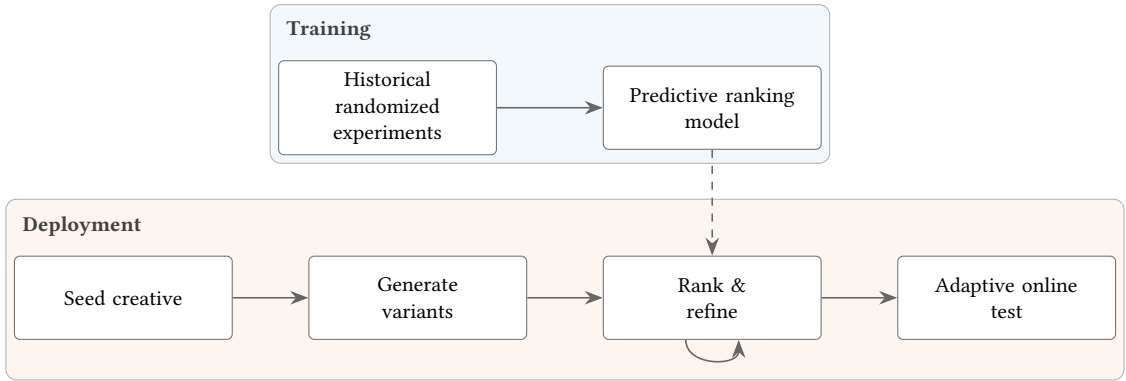

Given one or more seed creatives (e.g. written by a human), we use the ranking model to guide a generative model in generating and selecting a candidate slate \(S = \{y_1,\ldots,y_K\}\), run an online experiment with sample size \(N\), and select a final creative \(\hat{y}\) for deployment. Our workflow is summarized in Figure~\ref{fig:workflow}.

In the main study, the slate comprises 10 human-written creatives and 40 AI-refined variants. The AI-refined arms are produced by prompting a generative model with the human seed creatives and refining them under the offline ranking model (\S\ref{sec:method}).

At deployment time, the context \(x_0\) is fixed. To reduce notation, we write \(\mu(y)\) for \(\mu(x_0,y)\) in the evaluation metrics below.

\subsection{Evaluation Metrics}

We evaluate the workflow using three quantities.

\textit{Candidate-set recall.}
The offline and generative stages should produce a candidate set that contains strong creatives. Let \(T_m\) denote the realized top-\(m\) creatives according to their realized engagement rates or high-precision estimates thereof, and let \(R_k\) denote the top-\(k\) creatives according to the offline ranking model. We use $\mathrm{Recall@}k=\frac{|T_m \cap R_k|}{|T_m|}$ as a measure of whether the offline model surfaces high-performing candidates, even if its top-ranked candidate is not the realized best. We report multiple values, \(m \in \{5, 8, 10\}\). We also use threshold-based winner sets---arms exceeding the mean or the best human-generated arm---with recall defined analogously.

\textit{Selected-arm quality.}
The final decision quality is the engagement rate of the selected creative, $\mu(\hat{y})$. In the field experiment results, we report lift relative to two benchmarks: the best human-generated creative and the mean human-generated creative.

\textit{Online exploration regret.}
For a fixed slate \(S\), define the best creative in the slate as
\[
\mu_S^* = \max_{y \in S} \mu(y).
\]
If the online experiment assigns impression \(t\) to creative \(y_t\), the cumulative exploration regret is
\[
\mathrm{Regret}(S)
=
\sum_{t=1}^{N}
\left(\mu_S^* - \mu(y_t)\right).
\]
This measures the opportunity cost of screening the candidate slate online. We report regret per impression, which is comparable across slates of different sizes (Figure~\ref{fig:exploration_cost}).

For a slate of size \(K\), the chance of deploying a high-performing winner depends first on whether the winner is in the slate, i.e. Recall@K, and then on whether the online stage can identify it, i.e. the probability the bandit identifies the best arm. The simulation in \S\ref{sec:adaptive} shows that the second factor approaches one at modest experiment sizes for the wider slate, making candidate-set recall the binding offline requirement.

\section{Method: Generate Offline, Select Online}
\label{sec:method}

\subsection{Learning to Rank Improvements}

Following \citep{angelopoulos2024causal}, we train a ranking model using within-experiment comparisons from historical A/B tests. The use of within-experiment variation helps control for campaign-level factors that shift baseline performance. For an experiment with multiple arms, suppose creatives can be ordered by realized performance. If \(L\), \(M\), and \(H\) denote lower-, middle-, and higher-performing creatives within the same experiment, we construct comparisons such as $[M; H] \succ [M; L]$ that train the model to prefer changing the creative toward $H$ over changing the creative toward $L$ given the same reference creative $M$.

The ranking model can be represented in Bradley--Terry form:
\[
\mathbb{P}_{\theta}(H \succ L | M)
=
\sigma\left(s_{\theta}(H; M) - s_{\theta}(L; M)\right),
\]
where \(s_{\theta}\) is a learned scoring function and \(\sigma(\cdot)\) is the logistic function. At inference time, given base creative $y_0$ and proposed alternatives $\{y_1, \dots, y_K\}$, the alternatives can be ordered by $s_\theta(y_i; y_0)$. 
\citet{angelopoulos2024causal} use this ranking model to re-rank generated improvements; here we use the same ranker component to steer inference-time candidate generation and construct a slate that is then screened by a new online adaptive experiment.

\subsection{Ranker-Guided Candidate Generation}

At inference time, we use the ranking model to steer a frozen generative model. Given a human-written seed creative, the system first prompts the generator for multiple alternatives. The offline ranker scores the alternatives, producing an ordering but not a deployment decision. The generator is then asked to edit the current best candidate using the ranker-induced ordering as the optimization signal, and it produces a second batch of refined alternatives. The candidates from both iterations are pooled, deduplicated, filtered, and re-ranked before online testing. Importantly, we do not require task-specific fine-tuning of the generator.

Intuitively, our procedure is a one-step gradient ascent following TextGrad: generate candidates in parallel, rank them with an external critic, use the ranking to propose local edits, and select from the pooled candidates. It is analogous to textual-gradient and Best-of-$N$ language-space search \citep{yuksekgonul2024textgrad,kang2025textbo}, but the critic in our system is a ranking model trained from randomized advertising experiments rather than an LLM judge or environment feedback.

The goal is not to deploy the top offline-ranked creative directly. The goal is to enrich the slate with promising alternatives that can be evaluated online. In the field deployment, the candidate slate also passed human review before launch. The recall and precision results below therefore characterize the ranking model on the deployed slate; slate composition reflects generation, offline ranking, filtering, and review together.

\subsection{Adaptive Online Testing}

The online stage turns the slate of candidates into a deployment decision. The candidate slate is evaluated in a batched adaptive experiment, where the first batch uses uniform allocation across candidates. Subsequent batches update posterior beliefs over arm engagement rates and reallocate impressions toward candidates that are likely to be optimal or likely to challenge the current leader.

In our field deployment, we use Thompson sampling. Concretely, we model each arm's reward with a Beta--Bernoulli posterior, begin with a uniform first batch, and in each subsequent batch allocate impressions in proportion to each arm's posterior probability of being optimal.

The online experiment serves two purposes. First, it validates candidates generated by the offline model using actual user feedback. Second, it reduces the opportunity cost of testing by reallocating traffic away from weak candidates.

\section{System Deployment Considerations}
Our deployed system comprises several interconnected pieces of infrastructure.

\textit{Ranking model training.} For the main field deployment, the ranker was trained on historical experiments specific to the advertiser account running the optimization. The model consists of a value head attached to a Llama-3.2-1B \citep{grattafiori2024llama} backbone and is trained with LoRA \citep{hu2022lora} using the TRL library \citep{vonwerra2020trl}.

An important consideration for training the ranking model is whether the historical dataset used should be constructed using only ads from that particular advertiser, or whether experimental results should be aggregated across multiple advertisers. There is a bias-variance tradeoff: pooling reduces variance though may introduce bias. We leave an empirical evaluation of this tradeoff to future work.
In the offline evaluation of \S\ref{sec:offline-diagnostic}, we used a larger ranker trained on a broad cross-advertiser corpus, with an account-level train/validation split.

\textit{Guided candidate generation.} We guide an off-the-shelf LLM to generate candidates with the predictive ranking model. Given a seed creative, we make three calls to the LLM and two calls to the ranking model as follows:
\begin{enumerate}
    \item Initial generation: we prompt the LLM to generate 5 variants that it thinks will outperform the given creative.
    \item Ranking: we use the predictive model to rank the 5 variants.
    \item Refinement: we put the rank-ordered list of creatives in the prompt of the LLM and ask it to generate 5 new variants that it thinks will outperform the current top performer conditional on these results. We use a two-step prompt following TextGrad \cite{yuksekgonul2024textgrad}.
    \item Re-ranking: we use the predictive model to re-rank all 10 variants.
\end{enumerate}
We repeat this process in parallel multiple times and present the top variant of each run for human review. Specific prompts are in Appendix \ref{sec:prompts}.

\textit{Adaptive online testing.} The top candidates from guided candidate generation are piped into Ax, Meta's open-source platform for adaptive experimentation \citep{olson2025ax}. Ax is fully integrated with Meta's online experimentation platform, and includes algorithms for batch bandit experiments.

\section{Offline Stage Evaluation: Does Ranker-Guided Refinement Improve Candidates?}
\label{sec:offline-diagnostic}

Before turning to the field experiments, we study the extent to which the ranker-guided refinement step moves generated candidates in the intended direction. For this study, we train two ranking models: one on a training set of historical experiments, and the other on a disjoint holdout set. The training set model is used to guide candidate refinement, and then the refined candidates are evaluated under the holdout model, which for this offline analysis serves as a proxy for real user traffic. This study checks whether the refinement step improves candidates under an independent scorer, rather than under the model used to guide the optimization. For this exercise, we use a broad corpus of experiments rather than the account-specific ranker used in the main deployment. We score each candidate with both the \emph{in-sample} model that guided generation, and the \emph{out-of-sample} model trained on the holdout set.

\begin{figure}[h]
\centering
\includegraphics[width=.95\linewidth]{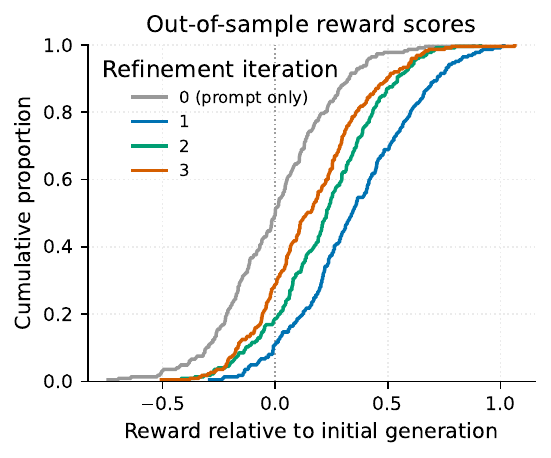}
\caption{Offline diagnostic: empirical CDF of candidate reward relative to the initial (iteration-0) generation, by refinement iteration, scored by an out-of-sample reward model trained on a held-out split. One refinement iteration shifts the distribution rightward of the iteration-0 (prompt-only) baseline (dashed line at $0$); further iterations partly regress, evidence that the gain is real but that one iteration is the right operating point. Reward is the latent Bradley--Terry score, so only differences are meaningful. The in-sample counterpart is Figure~\ref{fig:offline-diagnostic-insample} in the appendix.}
\label{fig:offline-diagnostic}
\end{figure}

Starting from seed ad text, we generate an initial batch of alternatives and then apply up to three ranker-guided refinement iterations, denoting the initial generation as iteration~0. We measure each candidate's reward score relative to this initial generation. One refinement iteration shifts the score distribution clearly to the right (Figure~\ref{fig:offline-diagnostic}): under the out-of-sample model the median one-iteration candidate reaches the 91st percentile of the prompt-only distribution, indicating that the shift is not merely overfitting to the model that guided generation. Returns from further iterations are negative, which indicates overfitting to the in-sample reward and motivates the single refinement iteration used in the field deployment. Table~\ref{tab:textgrad-gains} reports each iteration's mean reward gain over the initial generation: one iteration adds about $0.3$ under the in-sample model---equivalent, under the Bradley--Terry model, to a 57\% win rate of the refined text over the initial generation---with little gain thereafter.


\begin{table}[t]
\centering
\input{figures/offline_diagnostic_gains.tex}
\caption{Mean reward gain over the initial (iteration-0) generation, by refinement iteration, under the in-sample and out-of-sample reward models. Latent Bradley-Terry scores are shown, so only differences are meaningful. One round of refinement increases rewards, while deeper rounds show overfitting.}
\label{tab:textgrad-gains}
\end{table}

\section{Evidence and Lessons Learned from the Field}
\label{sec:field}

We evaluate the workflow in three large-scale optimizations run on our deployed system. These optimizations were conducted through different channels and had different target metrics. The main study is a 50-arm promotional email optimized for top-of-funnel engagement, with 10 human-written creatives and 40 AI-refined variants. We also refer to the human creatives as ``business-as-usual'' (BAU). There are 5.1 million recipients split into 3 batches. Two supporting experiments used the same workflow in different settings---a push notification and a second promotional email, comprising 13 and 32 arms---and produced qualitatively similar results. We focus on the 50-arm study for detailed analysis because it has the richest slate, and summarize the two supporting experiments in Section~\ref{sec:supporting}.

\subsection{AI-Refined Candidates Improve the Candidate Slate}

\begin{figure}[h]
\centering
\includegraphics[width=\linewidth]{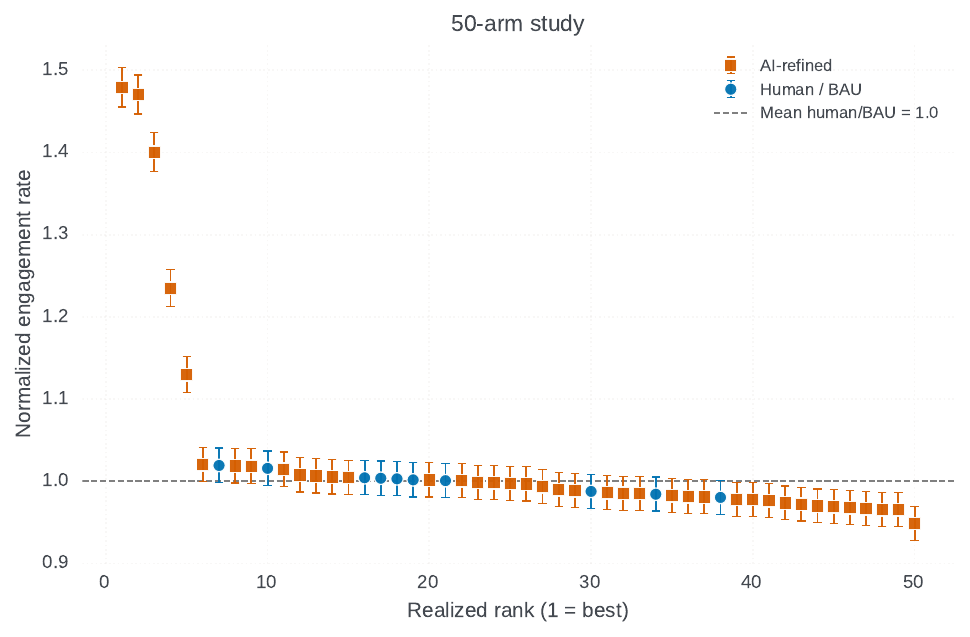}
\caption{Realized performance of human-generated and AI-refined creatives in the main study, ranked best to worst. Units are normalized so that the mean of the human-generated arms is 1.0. The best AI-refined creative exceeds the best human-generated creative by 45.1\%. Error bars are Wilson 95\% intervals computed from each arm's realized impression count from the first batch.}
\label{fig:rank-plot}
\end{figure}

The AI-refined candidates clearly improve the upper tail of the candidate slate. In the field experiment, the six top-performing creatives are all AI-refined, seven of the top eight are AI-refined (Table~\ref{tab:selected_arm_quality}), and the best AI-refined creative achieves a 45.1\% higher open rate than the best human-generated creative (Figure~\ref{fig:rank-plot}). This is the first requirement of the offline stage---the generated slate contains candidates worth finding.

\begin{table}[h]
\centering
\begin{tabular}{cccc}
\toprule
$m$ & AI-refined & Human & AI fraction \\
\midrule
3  & 3  & 0 & 1.00 \\
5  & 5  & 0 & 1.00 \\
8  & 7  & 1 & 0.88 \\
10 & 8  & 2 & 0.80 \\
15 & 13 & 2 & 0.87 \\
\bottomrule
\end{tabular}
\caption{Realized top-$m$ composition (main study): counts of AI-refined and human arms among the top $m$ by realized open rate.}
\label{tab:selected_arm_quality}
\end{table}

\subsection{The Offline Model Has Useful Recall but Limited Top-Rank Precision}
\label{sec:recall}

We evaluate the offline ranking model as a candidate-set generator under two winner definitions: arms that exceed the \emph{mean} human-generated arm and arms that exceed the \emph{best} human-generated arm (Table~\ref{tab:recall}). The offline model cannot reliably pick the single best creative, but its recall rises steeply with $k$: screening the offline top 30 retains 73\% of above-mean winners and 83\% of above-max winners, while pruning 20 of the 50 arms. Figure~\ref{fig:recall} (appendix) plots Recall@$k$ across both definitions and three realized-top-$m$ cutoffs.

\begin{table}[h]
    \centering
\begin{tabular}{c cc cc}
\toprule
 & \multicolumn{2}{c}{Above mean human} & \multicolumn{2}{c}{Above best human} \\
\cmidrule(lr){2-3} \cmidrule(lr){4-5}
$k$ & Precision & Recall & Precision & Recall \\
\midrule
1  & 0.00 & 0.00 & 0.00 & 0.00 \\
3  & 0.67 & 0.09 & 0.00 & 0.00 \\
5  & 0.60 & 0.14 & 0.00 & 0.00 \\
10 & 0.30 & 0.14 & 0.00 & 0.00 \\
15 & 0.47 & 0.32 & 0.13 & 0.33 \\
20 & 0.50 & 0.45 & 0.20 & 0.67 \\
25 & 0.52 & 0.59 & 0.16 & 0.67 \\
30 & 0.53 & 0.73 & 0.17 & 0.83 \\
\bottomrule
\end{tabular}
    \caption{Precision and recall of the offline ranking model at identifying winners, under two winner definitions: arms exceeding the \emph{mean} human-generated arm ($22$ winners) and arms exceeding the \emph{best} human-generated arm ($6$ winners). Recall rises steeply with $k$ under both; surfacing the rare above-best ``super-variants'' requires screening many arms---expensive under a fixed split test, but cheaper under adaptive allocation (\S\ref{sec:adaptive}).}
    \label{tab:recall}
\end{table}

The offline model's single highest-ranked creative finishes 43rd of 50 in realized open rate, below the average human arm, so its most confident pick would be a poor deployment choice. Conversely, the realized best creative sits at offline rank 19 of 50---outside the offline top 10 but inside the top 30---so it is surfaced by a slate wide enough to screen, not by a single confident pick. The model is therefore well-suited as a slate-construction tool, since once the slate is wide enough, recall is high enough for online testing to find a strong arm. This also cautions against aggressive pruning: the offline top-10 slate contains no AI-refined arm that beats the best human creative, while the top-30 slate contains five of the six such arms.

\subsection{Adaptive Testing Increases Online Exploration Efficiency}
\label{sec:adaptive}

We assess the online stage with a semi-synthetic simulation calibrated to the field run. Each arm's true open rate is set to its observed point estimate $\mu_a$ from the deployment; per-impression rewards are Bernoulli($\mu_a$). The $\mu_a$ are point estimates with sampling error, so this exercise is a controlled illustration of allocation dynamics under a plug-in estimate of rewards. We form candidate slates from the offline top-$K$ AI-refined arms ($K \in \{10, 30\}$) together with all 10 human arms, and compare two allocation policies: uniform allocation and Thompson sampling. Each configuration runs for three batches (the first uniform) and is averaged over 1{,}000 Monte Carlo replications. This isolates the effect of the allocation rule from the candidate set, holding the field's plug-in rewards fixed.

Two findings emerge, corresponding to the two things online traffic must do: identify a winner, and earn reward while doing so.

\begin{figure}[t]
\centering
\includegraphics[width=\linewidth]{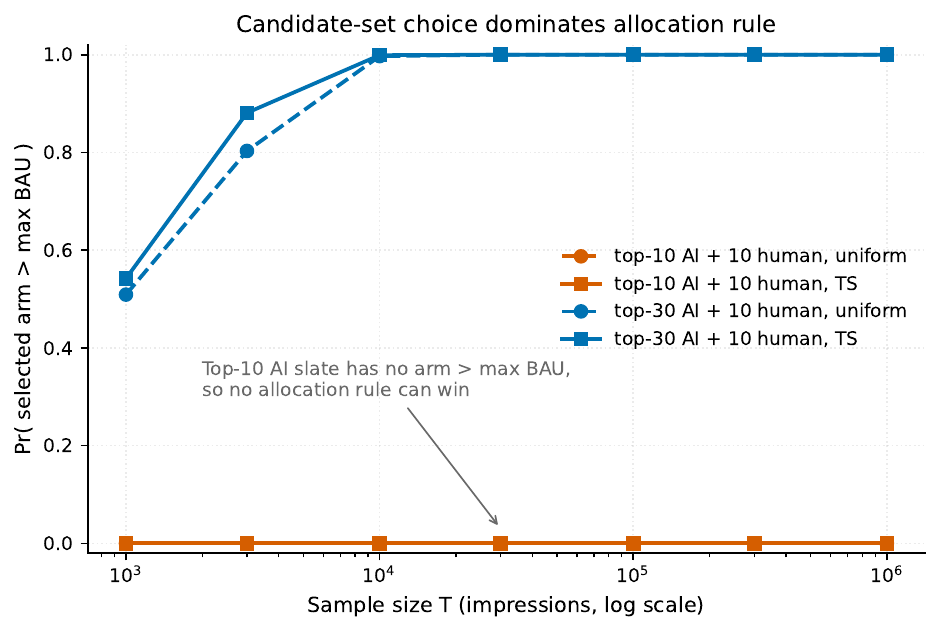}
\caption{Final-arm identification versus sample size $T$: probability the selected arm beats the best human arm. At $K=10$ the slate contains no arm that beats the best human (Recall@10 $=0$), so the line is pinned at zero for both policies; at $K=30$ both policies recover a winner by $T\approx 10^4$. Candidate-set choice (orange vs.\ blue) dominates allocation rule (uniform vs.\ Thompson sampling).}
\label{fig:identification}
\end{figure}

\textit{Identification.} Figure~\ref{fig:identification} reports the probability that the final selected arm beats the best human arm. At $K=10$, none of the offline top-10 AI arms exceed the best human arm (Table~\ref{tab:recall}, Recall@10 $=0$), so no slate arm can beat the best human and this probability is zero for every sample size and both policies. At $K=30$, the slate contains arms that beat the best human, and both policies recover one with high probability by $T \approx 10^4$. Notably, the allocation rule barely affects identification: uniform and Thompson sampling track each other closely throughout, while widening the candidate set from 10 to 30 moves the line from zero to one.

\begin{figure}[t]
\centering
\includegraphics[width=.9\linewidth]{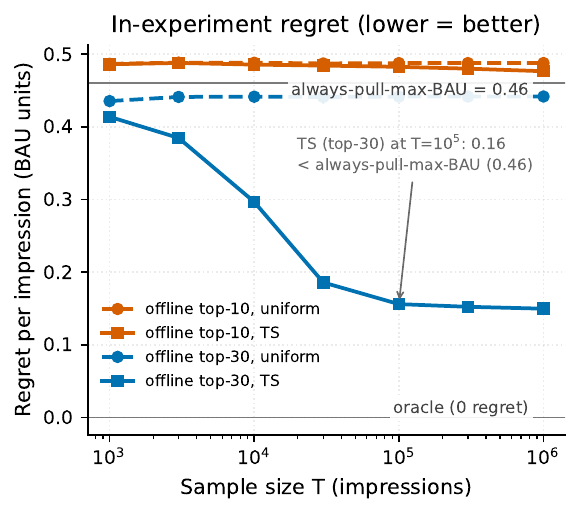}
\caption{In-experiment regret per impression ($\mu^* - $ reward), in units of the mean human/BAU open rate (BAU mean $=1$), with reference lines for the oracle ($0$ regret) and the always-pull-max-BAU regret. At $K=30$, Thompson sampling closes 66\% of the always-pull-max-BAU regret at $T=10^5$. The corresponding reward per impression is Figure~\ref{fig:exploration_cost_reward} in the appendix.}
\label{fig:exploration_cost}
\end{figure}

\textit{Exploration regret.} The policies differ sharply in what the experiment earns en route. Figure~\ref{fig:exploration_cost} reports in-experiment regret per impression, with reward per impression in Figure~\ref{fig:exploration_cost_reward} (appendix). At $K=30$ and $T=10^5$, Thompson sampling reduces in-experiment regret by 66\% relative to always allocating to the best human arm (Table~\ref{tab:sim_summary_focal}, appendix). The benefit grows with slate size, since uniform allocation spends a larger share of traffic on weak arms as $K$ increases.

Because the field deployment was intentionally large, both adaptive and fixed balanced allocation collect enough data to identify a strong final arm at the full sample size. The observed benefit of adaptivity at this scale is therefore primarily reduced online exploration cost rather than improved final selection.

\subsection{Additional Field Studies}
\label{sec:supporting}

The same workflow was deployed in two further field studies: a push-notification campaign (13 arms: 10 AI-refined, 3 human, $N=2.7M$) and a second promotional email campaign (32 arms: 25 AI-refined, 7 human, $N=2.1M$). To avoid confounding from the bandit's time-varying allocation, we evaluate each arm on its first, equal-allocation batch.

\begin{figure*}[h]
\centering
\includegraphics[width=.8\linewidth]{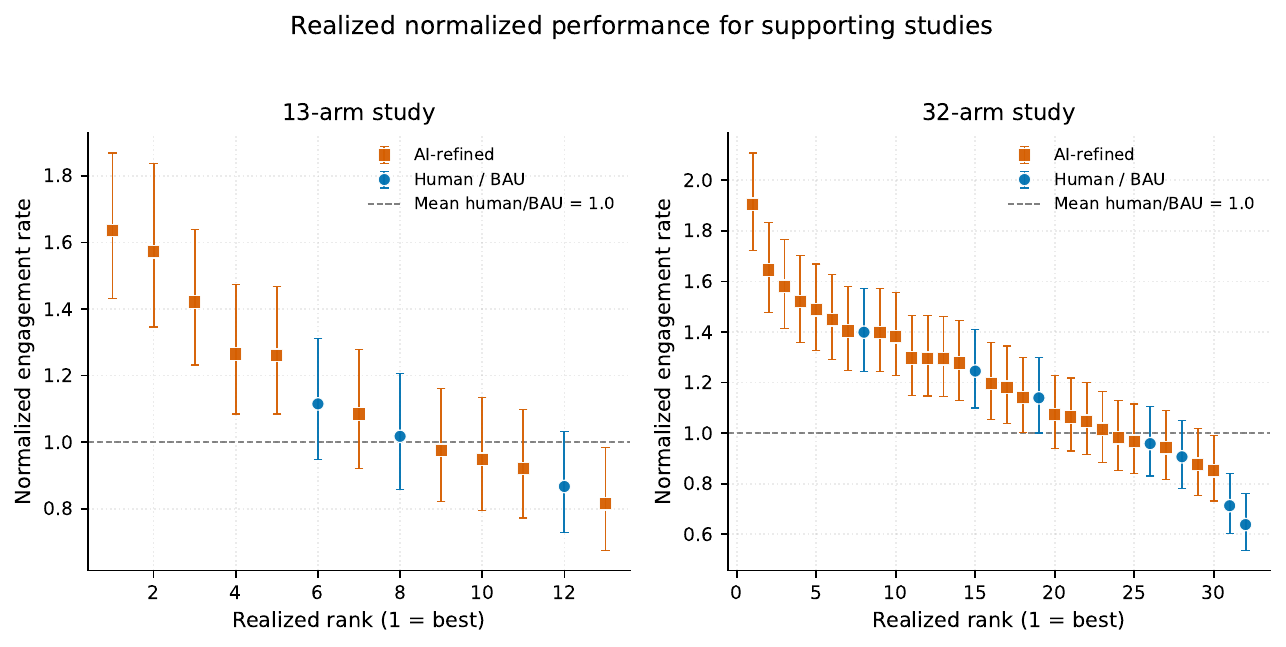}
\caption{Realized per-arm performance for the two supporting field experiments, ranked best to worst and normalized so the mean human/BAU arm equals 1.0 within each study. AI-refined arms (squares) concentrate in the upper tail. Error bars are Wilson 95\% intervals.}
\label{fig:cross-study}
\end{figure*}

The main-study conclusions are consistently found in both (Figure~\ref{fig:cross-study}): AI-refined arms concentrate in the upper tail, and the best AI-refined creative exceeds the best human creative by 46.7\% in the push campaign and 36.2\% in the email campaign, comparable to the 45.1\% improvement in the main study.

\section{Discussion and Future Directions}

In high-dimensional spaces, it can be difficult to train a predictive model that can accurately and reliably select the top creative. Our results have demonstrated that instead we can use it to help construct a slate that contains strong candidates. Online adaptive testing can then convert candidate-set recall into a high quality selected-arm while improving the efficiency of screening.

This distinction matters for how such systems should be evaluated. Offline top-1 accuracy is the wrong primary metric when the predictive model is not the deployment rule. The relevant offline question is whether a screenable slate contains high-performing candidates. The relevant online questions are whether the experiment identifies one of those candidates, and how much experiment traffic is given to weak candidates in the process. Our field deployment and calibrated simulations evaluate all three margins.

A limitation of our work is that it is data-dependent by design. It requires a historical corpus of randomized experiments over creative units that are sufficiently related to the deployment setting, and it requires online feedback that arrives quickly enough to screen a moderately large slate. We therefore do not claim that the specific ranking model studied here transfers unchanged across platforms, advertisers, objectives, or content formats that have little historical experimentation. The online stage mitigates but does not eliminate this dependence: it can identify strong arms once they are in the slate, but it cannot recover a winner that the offline generation and filtering process never surfaces. However, the broader lesson is portable: when the predictive model feeds an online experiment rather than a deployment decision, candidate-set recall matters more than top-1 prediction accuracy.

Our field evidence and supporting experiments suggest this condition can hold in large-scale advertising systems with repeated randomized creative tests, but they need not hold for smaller advertisers, long-lag conversion objectives, heavily regulated content, or settings where historical experiments cover a narrow creative distribution. Understanding how recall scales with the size and diversity of historical experiments is an important direction for future work.

We report engagement outcomes but do not report downstream conversion outcomes, which are noisier at the arm level and observed with higher latency. Future work can study how ranker-guided generation and adaptive screening affect downstream business outcomes, the optimal scale of candidate generation, and the margin of which campaigns should be experimentally optimized.

Generative models change the scale of creative experimentation. They make it easier to produce plausible candidates, but they also make the evaluation problem harder by expanding the set of arms that could be tested. The evidence here suggests that the value of generative AI in creative optimization comes from coupling generation with experimentation: predictive models help produce a slate with upper-tail candidates, and online tests remain necessary to identify which of those candidates should be deployed.

\bibliographystyle{ACM-Reference-Format}
\bibliography{refs}

\appendix
\section{Prompts}
\label{sec:prompts}

Below are the raw prompts used in the variant generation.

\begin{promptbox}[label={prompt:initial}]{Initial generation}
You are an expert copywriter. For the following creative:

{base_text}

Generate 5 alternative versions that would perform better on {engagement_metric}. Vary your approaches -- try different angles, tones, hooks, and structures. Keep approximately the same length. Return ONLY a JSON array of exactly 5 strings.
\end{promptbox}

TextGrad consists of two steps: a critique and an iteration. Prompts for both are below.

\begin{promptbox}{Critique}
Below is a rank-ordered list of text creative candidates for the same purpose, ranked from best-performing to worst-performing according to a predictive model.

{ranked_list}

---

The original text was: "{base_text}"

Analyze the patterns:
- What makes the top-ranked ones better than the bottom-ranked ones?
- What specific word choices, structures, or angles drive higher scores?
- What should be avoided based on the lower-ranked examples?

Give specific, actionable suggestions for further improvement. Be concise. Only give critiques and suggestions, don't give actual alternative texts.
\end{promptbox}

\begin{promptbox}{Iterate}
You are an expert copywriter editing a text creative to maximize {engagement_metric}.

Original text: "{base_text}"
Current best-performing version: "{top_ranked}"

Use these critiques from past results to help you revise the text:
---
{critique}
---

Generate 5 new alternatives that incorporate these insights and push the creative further. Try to beat the current best. Return ONLY a JSON array of exactly 5 strings.
\end{promptbox}

\section{Additional Results}
\label{sec:appendix}

\begin{figure}[H]
\centering
\includegraphics[width=.95\linewidth]{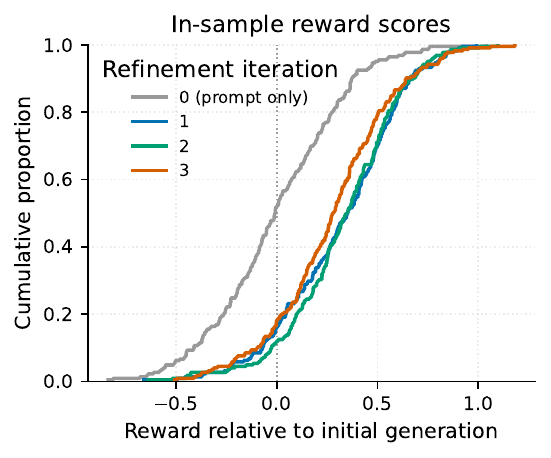}
\caption{Offline diagnostic, in-sample reward model: empirical CDF of candidate reward relative to the initial (iteration-0) generation, by refinement iteration, scored by the in-sample reward model (the split used to guide generation). As in the out-of-sample case (Figure~\ref{fig:offline-diagnostic}), one refinement iteration shifts the distribution rightward of the iteration-0 (prompt-only) baseline (dashed line at $0$), with little gain from further iterations. Reward is the latent Bradley--Terry score, so only differences are meaningful.}
\label{fig:offline-diagnostic-insample}
\end{figure}

\begin{figure}[H]
\centering
\includegraphics[width=.9\linewidth]{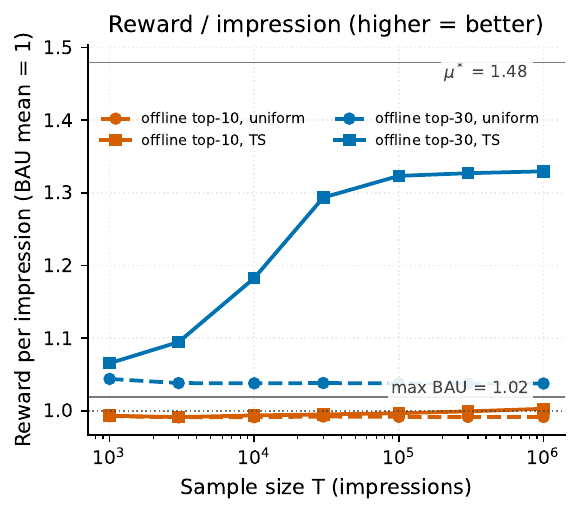}
\caption{Reward per impression, in units of the mean human/BAU open rate (BAU mean $=1$), with reference lines for $\mu^*$, the best human arm (max BAU), and the mean human arm. Reward counterpart to the in-experiment regret in Figure~\ref{fig:exploration_cost}: at $K=30$, Thompson sampling drives reward per impression above the best human arm as the budget grows, while uniform allocation and top-10 slates stay near baseline.}
\label{fig:exploration_cost_reward}
\end{figure}

\begin{figure}[H]
\centering
\includegraphics[width=\linewidth]{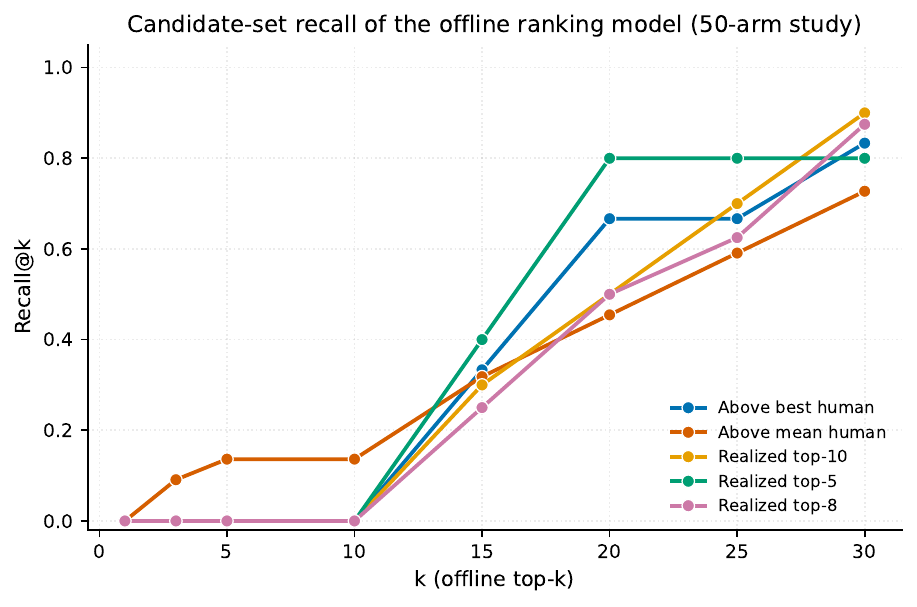}
\caption{Candidate-set recall of the offline ranking model (main study) under five winner definitions. Recall is near zero for small $k$ but rises sharply once $k \geq 15$.}
\label{fig:recall}
\end{figure}

This appendix collects supplementary results referenced in the main text: the in-sample counterpart of the refinement diagnostic (Figure~\ref{fig:offline-diagnostic-insample}), the reward-per-impression view of the exploration-cost simulation (Figure~\ref{fig:exploration_cost_reward}), candidate-set recall under all winner definitions (Figure~\ref{fig:recall}), and the focal simulation summary at $T=10^5$ (Table~\ref{tab:sim_summary_focal}).

\begin{table}[H]
\centering
\caption{Identification is decoupled from in-experiment cost at $T=100,000$. We vary the number of AI creatives to add to the slate of 10 human creatives. Within each slate, allocation policy does not affect final-arm selection (Pr beats max BAU) but affects what the experiment earns en route. Regret per impression is in normalized units.}
\label{tab:sim_summary_focal}
\begin{tabular}{llll}
\toprule
slate & policy & Pr($>$ max BAU) & Avg regret \\
\midrule
+ top-10 AI & Uniform & 0.0  & 0.488 \\
+ top-10 AI & Thompson & 0.0 & 0.483 \\
+ top-30 AI & Uniform & 1.0 & 0.442 \\
+ top-30 AI & Thompson & 1.0 & 0.156 \\
\bottomrule
\end{tabular}
\end{table}

\end{document}

%% file: figures/offline_diagnostic_gains.tex
\begin{tabular}{ccc}
\toprule
Iteration & In-sample & Out-of-sample \\
\midrule
0 & 0.000 & 0.000 \\
1 & 0.317 & 0.355 \\
2 & 0.330 & 0.221 \\
3 & 0.277 & 0.160 \\
\bottomrule
\end{tabular}